\documentclass{article}
\usepackage{ijcai19}

\usepackage{times}
\usepackage{helvet}  
\usepackage{courier}  
\usepackage{soul}
\usepackage{url}
\usepackage{array}
\usepackage{multirow}
\usepackage[hidelinks]{hyperref}
\usepackage[utf8]{inputenc}
\usepackage[small]{caption}
\usepackage{graphicx}
\usepackage{amssymb,amsmath}
\usepackage{epsfig,epstopdf}
\usepackage{subfigure}
\usepackage{booktabs}
\usepackage[ruled]{algorithm}
\usepackage{algorithmic}
\usepackage{hyperref}
\usepackage{comment}
\usepackage{placeins}
\usepackage{microtype}
\usepackage{balance}
\usepackage{color}
\usepackage[normalem]{ulem}
\usepackage{enumitem}
\usepackage{amsthm}
\newtheorem{myDef}{Definition}[section]

\title{ActiveHNE: Active Heterogeneous Network Embedding}
\author{
Xia Chen$^{1,3}$\and
Guoxian Yu$^1$\and
Jun Wang$^1$\and
Carlotta Domeniconi$^2$\and
Zhao Li$^{3}$\and
Xiangliang Zhang$^{4}$
\affiliations
$^1$College of Computer and Information Sciences, Southwest University, Chongqing, China\\
$^2$Department of Computer Science, George Mason University, VA, USA\\
$^3$Alibaba Group, Hangzhou, China\\
$^4$CEMSE, King Abdullah University of Science and Technology, Thuwal, SA\\
\emails
\{xchen, gxyu, kingjun\}@swu.edu.cn,
carlotta@cs.gmu.edu,
\{lizhao.lz\}@alibaba-inc.com,
xiangliang.zhang@kaust.edu.sa
}
\begin{document}
\maketitle
\begin{abstract}
Heterogeneous network embedding (HNE) is a challenging task due to the diverse node types and/or diverse relationships between nodes.
Existing HNE methods are typically  unsupervised.
To maximize the profit of utilizing the rare and valuable supervised information in HNEs, we develop a novel Active Heterogeneous Network Embedding (ActiveHNE) framework, which includes two components: Discriminative Heterogeneous Network Embedding (DHNE) and Active Query in Heterogeneous Networks (AQHN).
In DHNE, we introduce a novel semi-supervised heterogeneous network embedding method based on graph convolutional neural network. In AQHN, we first introduce three active selection strategies based on uncertainty and representativeness, and then derive a batch selection method that assembles these strategies using a multi-armed bandit mechanism. ActiveHNE aims at improving the performance of HNE by feeding the most valuable supervision obtained by AQHN into DHNE.
Experiments on public datasets demonstrate the effectiveness of ActiveHNE and its advantage on reducing the query cost.
\end{abstract}
%
%
\section{Introduction}
Networks are pervasive in a wide variety of real-world scenarios,  ranging from popular social networks, to citation graphs and gene regulatory networks.
Network embedding (NE), also known as network representation learning (NRL), enables us to capture the intrinsic information of the network data by embedding it into a low-dimensional space.
Effective NE approaches can facilitate downstream network analysis tasks, such as node classification, community discovery, and link prediction \cite{Cai2017A}.

Heterogeneous information networks (HINs), which involve diverse node types and/or diverse relationships between nodes, are ubiquitous in real-world scenarios \cite{Shi2017A}.
Although NE for homogeneous networks with single type of nodes and  single type of relationships has been extensively studied \cite{Tang2015LINE,Wang2016Structural,Cai2017A,Goyal2018survey}, the rich structure of HINs presents a major challenge for heterogeneous networks embedding (HNE), since nodes in different types should be treated differently (\textbf{Challenge 1}) \cite{Chang2015Heterogeneous,Fu2017HIN2Vec,Dong2017metapath2vec,Shi2018KDD,Chen2018KDD}.

Most of the current HNE approaches are unsupervised.
One can improve the performance of HNE by properly leveraging supervised information (\textbf{Challenge 2}).
However, label acquisition is usually difficult and expensive due to the involvement of human experts (\textbf{Challenge 3}).
For Challenge 3, active learning (AL), a technique widely used to acquire labels of nodes during learning,  can be adopted to save cost.
The selection of labeled data for model training can have significant influence on the prediction stage.
AL is expected to find the most valuable nodes to label with reduced query cost  \cite{settles2009active}.
However, since nodes in a heterogeneous network are not independently and identically distributed (\emph{non-i.i.d.}), but connected with links, AL with networks should account for data dependency.
In addition, for HINs, the different node types should also be considered.

\begin{figure}[t]
\centering
\includegraphics[width=8cm,height=4cm]{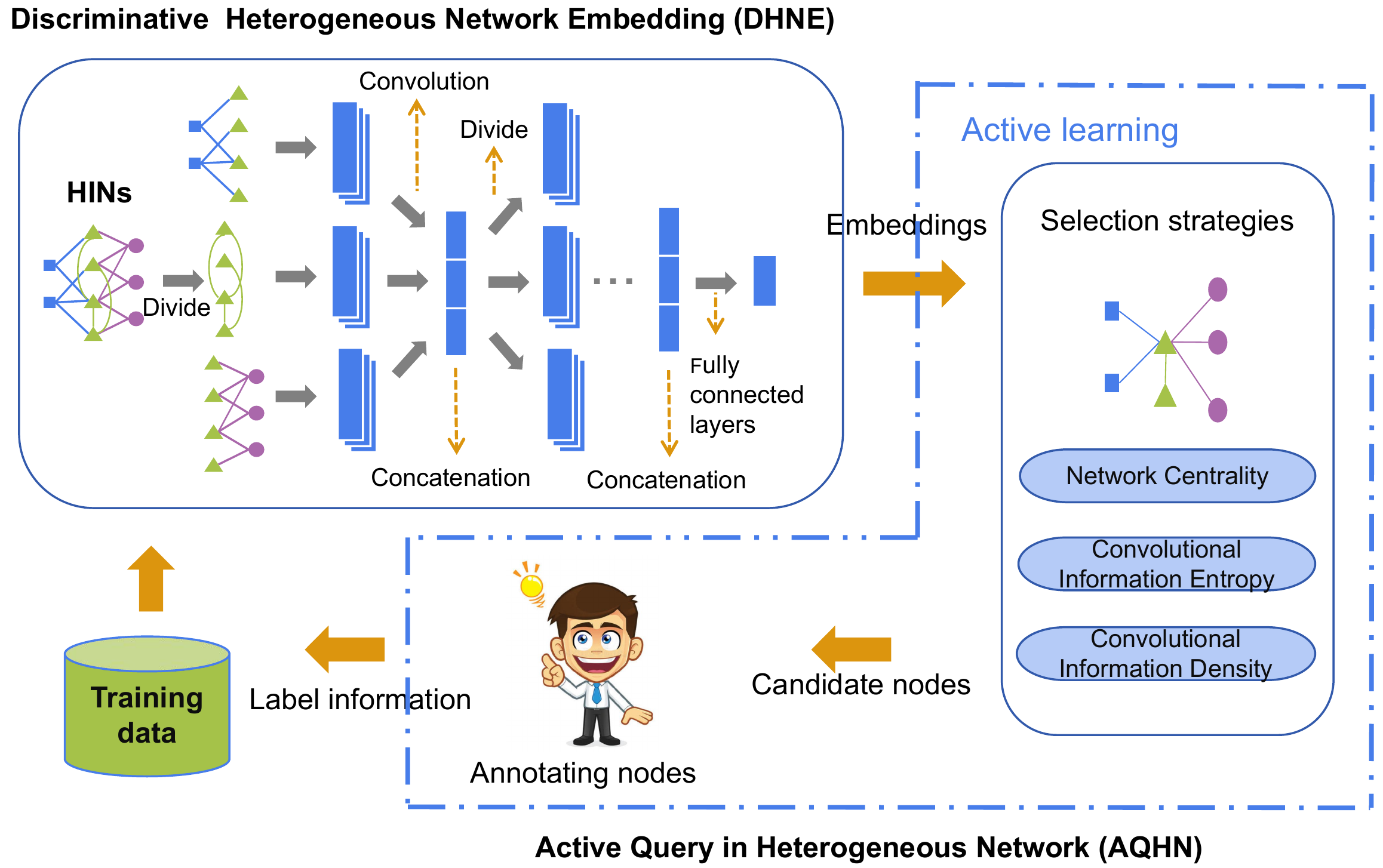}
\caption{The architecture of ActiveHNE. ActiveHNE consists of two components: Discriminative Heterogeneous Network Embedding (DHNE) and  Active Query in Heterogeneous Networks (AQHN). In each iteration, once a network embedding is obtained by DHNE,  AQHN  selects the most valuable nodes to be queried, and then updates DHNE with the new labels.}
\label{fig1}
\end{figure}

Based on the high efficiency of graph convolution networks (GCNs) \cite{Kipf2016Semi} in utilizing label information, we propose a novel Active Heterogeneous Network Embedding framework (called ActiveHNE) to address the above three challenges. ActiveHNE includes two components,  Discriminative Heterogeneous Network Embedding (DHNE) and Active Query in Heterogeneous Networks (AQHN), as illstrated in Figure \ref{fig1}. They are described below.
 \begin{itemize}
\item In DHNE, we introduce a semi-supervised discriminative heterogeneous network embedding method based on graph convolutional neural networks. Since different types of nodes and relationships should be treated differently, we first decompose the original HIN into homogeneous networks and bipartite networks. For each convolutional layer, DHNE separately learns the deep semantic meanings of nodes in each obtained network, and then concatenates the output vectors of each node from all networks.
\item In AQHN, besides the network centrality, we introduce two active selection strategies, namely convolutional information entropy  and convolutional information density for HINs with respect to uncertainty and representativeness. In particular, these strategies take advantage of the dependency among nodes and the heterogeneity of HINs by using local convolution, whose filter parameters are defined by the node importance (meassured by the number of node types of neighbors and the degree).
Then, we iteratively query the most valuable batch of nodes by combining the three strategies using the multi-armed bandit mechanism \cite{Sutton1998Reinforcement}.
\end{itemize}

This work makes the following contributions.  (i) We formalize the active heterogeneous network embedding problem, whose objective is to seek the most valuable nodes to query and to improve the performance of HNE using the queried labels. (ii) We present a novel heterogeneous graph convolutional neural network model for node embedding and node classification. (iii) Considering the data dependency among nodes and the heterogeneity of networks, we propose a new active learning method to select the most valuable nodes to label by leveraging local convolution and the multi-armed bandit mechanism. Experimental study on three real-world HINs demonstrate the effectiveness of ActiveHNE on embedding HINs, and on saving the query cost.

\section{Related Work}
Most of the previous approaches on HNE are unsupervised \cite{Shi2018KDD,Chang2015Heterogeneous,Gui2017Large,Xu2017Embedding}.
Recently, methods have been proposed to leverage meta-paths, either specified by users or derived from additional supervision \cite{Fu2017HIN2Vec,Dong2017metapath2vec,Shi2018AspEm}.
However, the choice of meta-paths strongly depends on the task at hands, thus limiting their ability of generalization \cite{Shi2018KDD}. In addition, they enrich the neighborhood of nodes, resulting in a denser network and in higher training costs \cite{Perozzi2014DeepWalk}.

Graph neural networks (GNNs) are another widely studied approach to leverage supervision \cite{Zhou2018Review}.
GNNs have the ability to extract multi-scale localized spatial features, and compose them to construct highly expressive representations.
Among all GNN approaches, graph convolution networks (GCNs) play a central role in capturing structural dependencies \cite{Wu2019Survey,Kipf2016Semi}.
A comprehensive survey of the literature shows that the majority of current GNNs are designed for homogeneous networks only.
GNNs are rarely explored for heterogeneous networks  \cite{Zhang2018DeepCollective}, and they are trained based on discretionary supervision.

One can improve the embedding performance by acquiring the labels of the most valuable nodes via AL. However, AL on  non-i.i.d. network data is seldom studied. In addition, the diversity of node  types in HINs makes the query criterion of AL even harder to design. Although attempts have been made to improve the embedding performance by incorporating AL,
they neither consider the dependence between nodes, nor the heterogeneity of networks \cite{Ye2017Active,Cai2017Active,Gao2018Active}.

\section{The ActiveHNE Framework}
In this section, we present our Active Heterogeneous Network Embedding framework, called ActiveHNE.
The architecture of ActiveHNE is given in Figure \ref{fig1}.
ActiveHNE consists of two components: Discriminative Heterogeneous Network Embedding (DHNE) and  Active Query in Heterogeneous Networks (AQHN), which are elaborated in the following subsections.

\subsection{Discriminative Heterogeneous Network Embedding (DHNE)}
\label{sec:31}
It's difficult to perform convolutions on networks due to the lack of an Euclidean representation space. In addition, HINs involve different types of nodes and relationships, each requiring its own processing, and further increasing the challenge of computing convolutions.
To address this issue, we first divide the original HIN into homogeneous networks and bipartite networks (the latter involving two types of nodes). After that, for each convolutional layer in a layer-wise convolutional neural network, we separately convolves and learns the deep semantic meanings of nodes in each obtained network, and then concatenates the output vectors of each node from all networks.

Let $\{\mathcal{G}_{t}|t=1,2,\cdots, T\}$ be the collection of obtained homogeneous networks and bipartite networks,
and $\{\mathbf{A}_t| t=1,2,\cdots, T\}$ denotes the adjacency matrices corresponding to $\{\mathcal{G}_{t}\}$.
The spectral graph convolution theorem defines the convolution in the Fourier domain based on the normalized graph Laplacian $\mathbf{L}_t={\mathbf{I}_t}-\mathbf{D}_t^{-\frac{1}{2}}\mathbf{A}_t\mathbf{D}_t^{-\frac{1}{2}}=\mathbf{D}_t^{-\frac{1}{2}}(\mathbf{D}_t-\mathbf{A}_t)\mathbf{D}_t^{-\frac{1}{2}}$, where ${\mathbf{I}_t}$ is the identity matrix and $\mathbf{D}_{t}=diag(\sum_{i}\mathbf{A}_{t}(i,j))$ denotes the degree matrix \cite{Kipf2016Semi,Chu2018Local}.

Since the nodes' degree distribution in an HIN may vary greatly, and the interaction between two connected nodes may be directed,  an asymmetric matrix $\mathbf{P}_t=\mathbf{D}_{t}^{-1}\mathbf{A}_{t}$, instead of the symmetric $\mathbf{L}_t$, is more suitable to define the Fourier domain.
$\mathbf{P}_t$ is the transition probability matrix.

In this paper, we separately convolve on each obtained network using the transition probability matrix $\mathbf{P}_t$ as Fourier basis.
Specifically, let $\mathbf{P}_t=\mathbf{\Phi}_t\mathbf{\Lambda}_t\mathbf{\Phi}_t^{-1}$, where $\mathbf{\Lambda}_t$ and $\mathbf{\Phi}_t$ are the eigenvector matrix and the diagonal matrix of eigenvalues of $\mathbf{P}_t$, respectively.
The convolution on each obtained network is defined as follows:
\begin{equation}
\begin{split}
g_{\mathbf{\theta}_t}\star \mathbf{X}_t=g_{\mathbf{\theta}_t}(\mathbf{P}_t) \mathbf{X}_t&=g_{\mathbf{\theta}_t}(\mathbf{\Phi}_t\mathbf{\Lambda}_t\mathbf{\Phi}_t^{-1}) \mathbf{X}_t\\
&=\mathbf{\Phi }_t g_{\mathbf{\theta}_t}(\mathbf{\Lambda}_t)\mathbf{\Phi}_t^{-1} \mathbf{X}_t
\label{eq2}
\end{split}
\end{equation}
where $\mathbf{X}_t \in \mathbb{R}^{N_t \times D}$ is the input signal of the network $\mathcal{G}_t$ ($N_t$ and $D$ denote the number of nodes and the number of features of each node in $\mathcal{G}_t$, respectively).
$g_{\mathbf{\theta}_t}\star \mathbf{X}_t$ gives the product of the signal $\mathbf{X}_t$ with a filter $g_{\mathbf{\theta}_t}$  in the graph Fourier domain, which denotes the output of  graph convolution.
$\mathbf{\Phi}_t^{-1} \mathbf{X}_t$ is the Fourier transform of signal $\mathbf{X}_t$. 
More details  about the spectral graph convolution in the Fourier domain can be found in \cite{Chu2018Local}.

To convolve the local neighbors of the target node, we define $g_{\mathbf{\theta}_t}(\mathbf{\Lambda}_t)$ as a polynomial filter up to $K$ order \cite{Defferrard2016Convolutional,Zhang2018DeepCollective} as follows:
\begin{equation}
g_{\mathbf{\theta}_t}(\mathbf{\Lambda}_t)= \sum_{k=1}^{K}{\mathbf{\theta}_t}_k\mathbf{\Lambda}^k
\label{eq3}
\end{equation}
where $\mathbf{\theta}_t\in \mathbb{R}^K$ is a vector of polynomial coefficients.
Thus, we have:
\begin{equation}
g_{\mathbf{\theta}_t}\star \mathbf{X}_t
=\mathbf{\Phi }_t(\sum_{k=1}^{K}{\mathbf{\theta}_t}_k\mathbf{\Lambda}_t^k)\mathbf{\Phi}_t^{-1} \mathbf{X}_t
=\sum_{k=1}^{K}{\mathbf{\theta}_t}_k\mathbf{P}_t^k\mathbf{X}_t
\label{eq4}
\end{equation}

From Eq. (\ref{eq4}), the convolution on $\mathcal{G}_t$ only depends on the nodes that are at most $K$ steps away from the target node.
In other words, the output signals after convolution operations are defined by a $K$-order approximation of localized spectral filters on networks.
The filter parameters ${\mathbf{\theta}_t}_k$ can be shared over the whole network $\mathcal{G}_t$.
Moreover, we generalize Eq. (\ref{eq4}) to $D \times d$ filters for feature maps, i.e., we map the original feature dimension $D$ to $d$.
Thus, the convolution operation on network $\mathcal{G}_t$ is formalized as follows:
\begin{equation}
\mathbf{H}_t=\sigma(\sum_{k=1}^{K}\mathbf{P}_t^k\mathbf{X}_t\mathbf{\Theta}_t)
\label{eq5}
\end{equation}
where $\mathbf{\Theta}_t \in \mathbb{R}^{D \times d}$ and $\mathbf{H}_t \in \mathbb{R}^{N_t \times d}$ denote the matrix of filter parameters (the trainable weight matrix) and the convolved signal matrix (output signals), respectively. We use $ReLU(\cdot)$ for $\sigma(\cdot)$ as the activation function.

So far, we have performed the convolutions separately on each individual network.
To leverage both the homologous and heterogeneous information of HINs for embedding, we then concatenate in order the vectors of the convoluted signals to obtain the final output signals for each node, according to the network it belongs to. For a node that is not an element of a network, we use a zero vector to represent the corresponding output signals.
Let $\mathbf{Z}_t$ denote the concatenated convoluted signals of nodes in $\mathcal{G}_t$, we define the layer-wise convolution on $\mathcal{G}_t$ as follows:
\begin{equation}
\mathbf{H}_t^{(l)}=\sigma(\sum_{k=1}^{K}\mathbf{P}_t^k\mathbf{Z}_t^{(l)}\mathbf{\Theta}_t^{(l)}), l=0,1,2,...
\label{eq6}
\end{equation}
where $\mathbf{Z}_t^{(l)}\in \mathbb{R}^{N_t \times Td^{(l-1)}}$, $\mathbf{\Theta}_t^{(l)} \in \mathbb{R}^{Td^{(l-1)} \times d^{(l)}}$, and $\mathbf{H}_t^{(l)} \in \mathbb{R}^{N_t \times d^{(l)}}$ denote the activations (input signals), the matrix of filter parameters (the trainable weight matrix), and the convolved signal matrix (output signals) in the $l$-th layer, respectively.
$d^{(l)}$ is the embedding dimension of the $l$-th layer, and $T$ is the number of networks.
Specifically, $\mathbf{Z}_t^{(0)}=\mathbf{X}_t$ and $\mathbf{\Theta}_t^{(0)} \in \mathbb{R}^{D \times d^{(1)}}$.
Eq. (\ref{eq6}) indicates the layer-wise propagation rule in layer-wise convolutional neural networks. Although we performed the convolutions separately on each individual network, both the homologous and heterogeneous information of HINs are used for the embedding thanks to the layer-wise concatenation operators.

After $\beta$ layers of convolutions and concatenations, we obtain the final output vectors of all nodes as $\mathbf{E} = \mathbf{Z}^{\beta} \in \mathbb{R}^{N\times Td^{(\beta)}}$. To obtain a discriminative embedding, we leverage supervision (i.e., label information) by adding a fully connected layer to predict the labels of nodes as follows:
\begin{equation}
\mathbf{F}=\sigma(\mathbf{E}\mathbf{\Theta}^{pre})
\label{eq7}
\end{equation}
where $\mathbf{\Theta}^{pre} \in \mathbb{R}^{Td^{(\beta)}\times C} $ is the hidden-to-output weight matrix). 
$\mathbf{F} \in \mathbb{R}^{N \times C}$, and
$\mathbf{F}_{ic}$ stores the probability that the $i$-th node belongs to class $c$.
The activation function $\sigma(\cdot)$ in the last layer is the softmax function, which is defined as $softmax(\mathbf{F}_{ic}) = \frac{exp(\mathbf{F}_{ic})}{\sum_{c'=1}^C exp(\mathbf{F}_{ic'})}$ 
Finally, the supervised loss function is defined as the cross-entropy error over all labeled nodes as follows:
\begin{equation}
loss=-\sum_{i=1}^{L} \sum_{c=1}^{C} \mathbf{Y}_{ic}ln\mathbf{F}_{ic}
\label{eq8}
\end{equation}
where $\mathbf{Y} \in \{0,1\}^{N\times C}$ stores the ground-truth labels of nodes. If the $i$-th node is associated with the $c$-th label, $\mathbf{Y}_{ic}=1$. Otherwise, $\mathbf{Y}_{ic}=0$.
The neural network weight parameters $\mathbf{\Theta}_t^{(l)}$ and $\mathbf{\Theta}^{pre}$ are optimized using gradient descent to minimize Eq. (\ref{eq8}).
As such, Eqs. (\ref{eq7}) and (\ref{eq8}) enable a semi-supervised model for discriminative node embedding.  The label of the $i$-th node can be predicted as $y_i = \arg\max_c \mathbf{F}_{ic}$.

\subsection{Active Query in Heterogeneous Networks (AQHN)}
In DHNE, we perform a semi-supervised heterogeneous network embedding, which requires the participation of label information. However, label acquisition is usually difficult and expensive due to the involvement of human experts. More importantly, different supervision may lead to different embedding performance.
To train a more effective DHNE, we propose an active query component, AQHN, to acquire the most valuable supervision within a given budget (e.g., the allowed maximum number of queries).

Uncertainty and representativeness are widely used criteria to select samples for query in AL. Uncertainty selects the sample that the current classification model is least certain, while representativeness selects the sample that can well represent the overall input patterns of unlabeled data. Empirical studies have shown that combining the two criteria can make more efficient selection strategies \cite{Huang2014Active}.
In the following, we first introduce three active selection strategies (Network Centrality, Convolutional Information Entropy, and Convolutional Information Density) for HINs based on uncertainty and representativeness.
Then, we proposed a novel method to combine these strategies to adaptively and iteratively select the most valuable batch of nodes to query, by leveraging the multi-armed bandit mechanism \cite{Sutton1998Reinforcement}.

\subsubsection{Selection Strategy}
\label{sec:Selection Strategy}

\indent \textbf{Network centrality (NC).} NC (e.g., degree centrality and closeness centrality)  \cite{Freeman1978Centrality} is an effective measure to evaluate the representativeness of nodes.
In this paper, we simply use degree centrality, which is defined as $\phi_{nc}(v_i)=|\mathcal{N}_i|$, to evaluate the centrality of nodes. $\mathcal{N}_i$ includes all the direct neighbors of $v_i$. Other measures of network centrality in HINs will be studied later.

Nodes in an HIN are non-i.i.d. and are connected by links, which reflect the dependency among nodes. Inspired by the idea of spectral graph convolution that defines the convoluted signal as a linear weighted sum of its neighbor signals, we propose two novel active strategies to select nodes for query in HINs based on a convolution of neighbors. We first define the convolution parameters (i.e., the weight parameters) and then the two selection strategies. Let $w_i =tanh(\frac{n_i}{N} + \frac{m_i}{V_T}) \in [0,1)$ to quantify the importance of node $v_i$. $tanh(\cdot)$ is the hyperbolic tangent function. Here $n_i$ and $m_i$ represent the number of neighbor nodes of $v_i$ and the number of node types of these neighbors. $N$ and $V_T$ are the total number of nodes and node types in the whole network, respectively.
A larger value of $n_i$ or $m_i$ implies that more complex information is conveyed by $v_i$, and thus $v_i$ may be more important to its neighbor nodes. In the following, we use $w_i$ as the weight parameters for convolving neighbors.

\textbf{Convolutional Information Entropy (CIE).}
Information Entropy (IE) is a widely used metric to evaluate uncertainty.
In this paper, we evaluate the uncertainty of node $v_i$ using CIE as follows:
\begin{equation}
\phi_{cie}(v_i)=\sum_{v_j \in \{v_i \bigcup \mathcal{N}_i\}} w_j(-\sum_{c=1}^{C}\mathbf{F}_{jc}\log \mathbf{F}_{jc})
\label{eq11}
\end{equation}
 The uncertainty of $v_i$ is a weighted sum of the uncertainties of its neighbors and itself.

\textbf{Convolutional Information Density (CID).}
The representativeness of nodes in the embedding space is also crucial to measure the value of nodes.
We apply $k$-means clustering  on the embedding to calculate the information density (ID) of nodes, due to its high efficiency. The number of clusters for $k$-means is simply set to the number of class labels. CID of $v_i$ based on its neighbors is quantified as follows:
\begin{equation}
\phi_{cid}(v_i)=\sum_{v_j \in \{v_i \bigcup \mathcal{N}_i\}} w_j \frac{1}{1+dis(\mathbf{E}_j,\varphi(v_j))}
\label{eq13}
\end{equation}
where $dis(\cdot)$ is the distance metric (i.e., Euclidean distance) in the embedding space, $\varphi(v_i)$ is the center vector of the cluster to which $v_i$ belongs. $\mathbf{E}_j$ is the embedding of the $j$-th node. $\varphi(v_j)$ and $\mathbf{E}_j$ belong to the same space.

The proposed CIE and CID measure the value (uncertainty or representativeness) of a node based on the node itself and its neighborhood nodes, while IE and ID are based on the node only. Since nodes in networks are connected by links, CIE and CID are more appropriate than IE and ID. We prove it in Section \ref{sec:Individual}.


\subsubsection{Multi-Armed Bandit for Active Node Selection}
We   select the most valuable nodes by leveraging the above three selection strategies.
In particular, we study the batch mode setting, in which we query $b$ nodes in each iteration.
First, we select top $b$ nodes with the highest  $\phi_{nc}$, $\phi_{cie}$, and $\phi_{cid}$ scores as the initial candidates of each selection strategy in each iteration, respectively.
To jointly select the most valuable $b$ nodes from all selection strategies, one can evaluate the score of each node by using the weighted sum of scores of each strategy, where the weights capture the importance of corresponding strategies.
Then, the problem of active node selection is transformed into the estimation of the importance of each strategy.
But the importance of each strategy is time-sensitive and thus difficult to be specified \cite{Cai2017Active,Gao2018Active}. We introduce a novel method to adaptively learn the dynamic weight parameters based on the multi-armed bandit mechanism.
The well-known multi-armed bandit (MAB) problem is a simplified version of the reinforcement learning problem 
\cite{Sutton1998Reinforcement}, which explores what a player should do given a  bandit machine with $\Lambda$ arms and a budget of iterations.
In each iteration, an agent plays one of the $\Lambda$ arms to receive a reward. The objective is to maximize the cumulative reward.
Combinatorial
MAB (CMAB) \cite{Chen2013Combinatorial}, an extension of MAB, allows to play multiple arms in each iteration.

Based on the idea of the CMAB, we can view each selection strategy as an arm, and approximate the importance of each strategy by estimating the expected reward (i.e., utility) of the corresponding arm.
Let $\mathcal{C}_r^\lambda$ be the initial candidate set of arm $\lambda$ in iteration $r$, and $\mathcal{Q}_r$ be the actually queried set of nodes in that iteration.
Intuitively, the actual reward of arm $\lambda$ can be defined as:
\begin{equation}
\mu_r(\lambda)=\psi(f_{\mathcal{L}_r \bigcup \mathcal{Q}_r^\lambda})-\psi(f_{\mathcal{L}_r})
\label{eq14}
\end{equation}
where $\mathcal{L}_r$ is the available labeled set of nodes in iteration $r$. $\mathcal{Q}_r^\lambda=\mathcal{C}_r^\lambda \bigcap \mathcal{Q}_r$ is the set of queried nodes that are dominated by arm $\lambda$ in iteration $r$.
$f_{\mathcal{L}_r}$ is the classifier trained on $\mathcal{L}_r$, and $\psi(f_{\mathcal{L}_r})$ is the classification performance of $f_{\mathcal{L}_r}$.
We observe that $\mu_r(\lambda)$ for the current iteration can't be computed since the ground-truth of $\mathcal{Q}_r^\lambda$ is unavailable.
The empirical reward is typically used to estimate the expected reward of arms. But computing the empirical $\mu_r(\lambda)$ of each arm in each iteration is very time-consuming; as such, we estimate the empirical reward of each arm using the local embedding changes of nodes caused by the arm.

We first define the local embedding changes caused by arm $\lambda$ in iteration $r$ as follows:
\begin{equation}
\Delta_r(\lambda) = \sum_{v_i \in \mathcal{Q}_r^{\lambda}}\sum_{v_j
\in \mathcal{N}(v_i)}dis(\mathbf{E}_j^r, \mathbf{E}_j^{r-1})
\label{eq15}
\end{equation}
where $dis(\cdot)$ is the distance metric (e.g., Euclidean distance), $\mathcal{N}(v_i)$ is the neighbors of $v_i$, and $\mathbf{E}_j^r$ is the node embedding of $v_j$ in iteration $r$.
Eq. (\ref{eq15}) measures the empirical reward of arm $\lambda$ in iteration $r$ using the local embedding changes of
nodes caused by the arm $\lambda$, which equates to the embedding changes of neighbor nodes of the nodes dominated by arm $\lambda$ (or $\mathcal{Q}_r^{\lambda}$).
This AL strategy aims to select nodes that result in the greatest change to the embeddings when their labels are available. The intuition is that one can view the magnitude of the resultant change of embeddings as the value of purchasing the labels. If this magnitude of change is small, then the labels do not provide much new information and has a low value.
To achieve a fair comparison and avoid bias,
the empirical reward of arm $\lambda$ in iteration $r$ is estimated as
$\hat{\mu}_r(\lambda)=\frac{\Delta_r(\lambda)}{\Delta_r(\bigcup_{\lambda=1}^{\Lambda}\lambda)}$, where $\Delta_r(\bigcup_{\lambda=1}^{\Lambda}\lambda)$ denotes the local embedding changes caused by all arms (or $\mathcal{Q}_r$).
Note that, in iteration $r$, $\Delta_r(\bigcup_{\lambda=1}^{\Lambda}\lambda) \leq \sum_{\lambda}^{\Lambda} \Delta_r(\lambda)$. The reason is that there may be overlap between different $\mathcal{Q}_r^{\lambda}$.
Due to the fact that the importance of each selection strategy changes over time,
we use the average of the last two empirical rewards to estimate the current expected reward as follows:
\begin{equation}
\bar{\mu}_r(\lambda)=\frac{\hat{\mu}_{r-2}(\lambda)+\hat{\mu}_{r-1}(\lambda)}{2}
\label{eq16}
\end{equation}

To mitigate the exploration-exploitation dilemma of CMAB, the combinatorial upper confidence bound algorithm \cite{Chen2013Combinatorial} estimates expected rewards based on the empirical rewards and the number of times an arm is explored.
In the same way, we adjust $\bar{\mu}_r(\lambda)$ as
$\tilde{\mu}_r(\lambda)=\overline{\mu}_r(\lambda)+\sqrt{\frac{3lnr}{2 n_{\lambda}}}$, where $n_{\lambda}$ denotes the total number of nodes queried by arm $\lambda$.
This adjustment can boost the expected reward of under-explored arms to avoid dismissing a potentially optimal strategy without sufficient evidences.

After this, to avoid selecting highly controversial nodes,
we estimate the expected reward of un-queried nodes $v_i \in \bigcup_{\lambda=1}^{\Lambda}\mathcal{C}_r^\lambda$ in iteration $r$  using the weighted Borda count as follows:
\begin{equation}
\tilde{\mu}^*_r(v_i)=\sum_{\lambda=1}^{\Lambda}\tilde{\mu}_r(\lambda)(b-rank_r^{\lambda}(v_i))
\label{eq17}
\end{equation}
where $rank_r^{\lambda}(v_i) \in [1,b]$ is the rank order of node $v_i$ in arm $\lambda$ in iteration $r$ (sorted in descending order of scores).
Finally, the top $b$ nodes (from $\bigcup_{\lambda=1}^{\Lambda}\mathcal{C}_r^\lambda$) with the highest $\tilde{\mu}^*_r(v_i)$ are selected as the query batch $\mathcal{Q}_r$ in iteration $r$.

\section{Experiments}
\subsection{Experimental Setup}
\textbf{Datasets}: we evaluate our ActiveHNE on three real-world HINs extracted from DBLP\footnote{\href{https://dblp.uni-trier.de/db/}{https://dblp.uni-trier.de/db/}}, Cora\footnote{\href{http://web.cs.ucla.edu/~yzsun/data/}{http://web.cs.ucla.edu/~yzsun/data/}}, and MovieLens\footnote{\href{https://grouplens.org/datasets/movielens/}{https://grouplens.org/datasets/movielens/}}.
The extracted DBLP consists of 14K papers, 20 conferences, 14K authors, and 9K terms, with a total of 171K links.
The extracted MovieLens includes 9.7K movies, 12K writers, 4.9K directors, 0.6K users, and 1.5K tags, with a total of 140K links. The extracted Cora has 25K authors, 19K papers, and 12K terms, with 146K links.

\textbf{Baselines}:
we compare ActiveHNE against the following state-of-the-art methods and a variant of ActiveHNE that randomly selects  nodes  to query (in a kind of naive AL setting):
\begin{itemize}
\item{\bf GCN} \cite{Kipf2016Semi}: a {\bf semi-supervised} network embedding 
model, with no consideration of networks heterogeneity. To adapt GCN in AL setting, nodes are randomly selected for query in each iteration (in  naive AL setting).
\item{\bf metapath2vec} \cite{Dong2017metapath2vec} and {\bf HHNE} \cite{HHNEAAAI18}: two {\bf unsupervised} HNE methods also adapted in the naive AL setting. 
\item{\bf AGE} \cite{Cai2017Active} and {\bf ANRMAB} \cite{Gao2018Active}: two {\bf active} network embedding methods
without considering the dependence between nodes and the heterogeneity of networks.
\item{\bf DHNE}: a variant of ActiveHNE that randomly selectsnodes to query in naive AL setting.
\end{itemize}

\textbf{Tasks}:
We evaluate the performance of network embedding using the Accuracy of node classification task.
For the DBLP dataset, we classify author, paper, and conference nodes into four research area: $\{$Data Mining, Database, Information Retrieval, Artificial Intelligence$\}$, with ground truth labels obtained in the same fashion as in \cite{Ming2011Ranking}. For the Cora dataset, after preprosessing, we classify paper nodes into ten research area 
$\{$Information Retrieval, Databases, Artificial Intelligence, Encryption and Compression, Operating Systems, Networking, Hardware and Architecture, Data Structures Algorithms and Theory, Programming, Human Computer Interaction$\}$. 
For MovieLens dataset, the movie nodes are classified into three genres $\{$Action, Romance, Thriller$\}$, with each node belonging to a single genre to satisfy the single label condition.

For the proposed DHNE and ActiveHNE, we simply set $K=1$ for comparative evaluation, and investigate $K$ in Section \ref{parameter_k}.
We train DHNE using a network with two convolutional layers and one fully connected layer as described in Section \ref{sec:31}, with a maximum of 200 epochs (training iterations) using Adam. The dimensionality of the two convolutional filters is 16 and $C$, respectively. We use an $L_2$ regularization factor for all the three layers. The remaining parameters are fixed  as in GCN \cite{Kipf2016Semi}.
For metapath2vec and HHNE, we apply the  commonly used meta-path schemes ``APA" and ``APCPA"  on DBLP and Cora, and we use ``DMTMD" and ``DMUMD" on MovieLens to guide metapath-based random walks. The walk length and the number of walks per node are set to 80 and 40 as in HHNE, respectively.

\begin{figure*}[h!t]
\centering
\hspace{-1em}
\subfigure[MovieLens]{\label{fig2a}\includegraphics[scale=0.4]{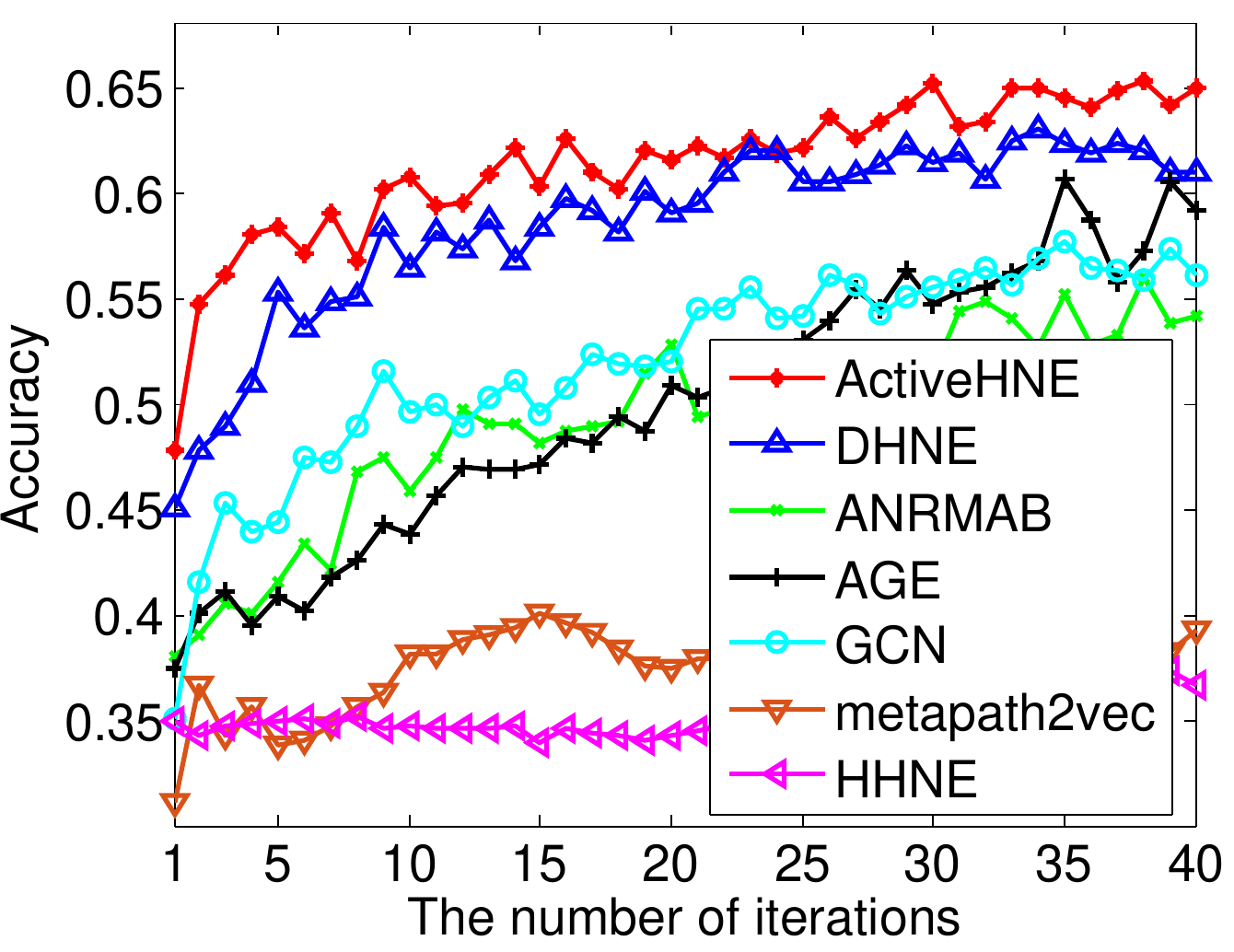}}
\subfigure[Cora]{\label{fig2b}\includegraphics[scale=0.4]{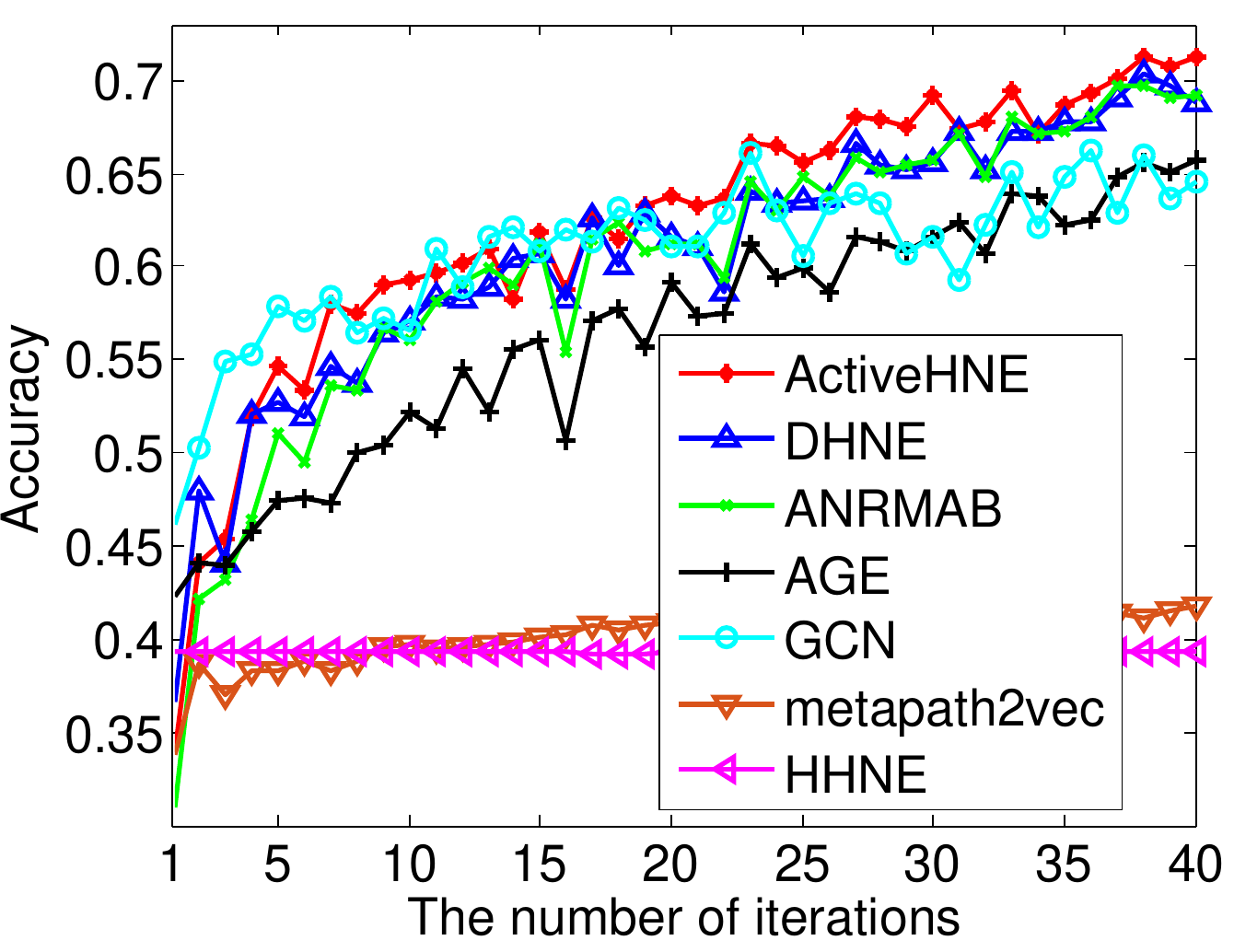}}
\subfigure[DBLP]{\label{fig2c}\includegraphics[scale=0.4]{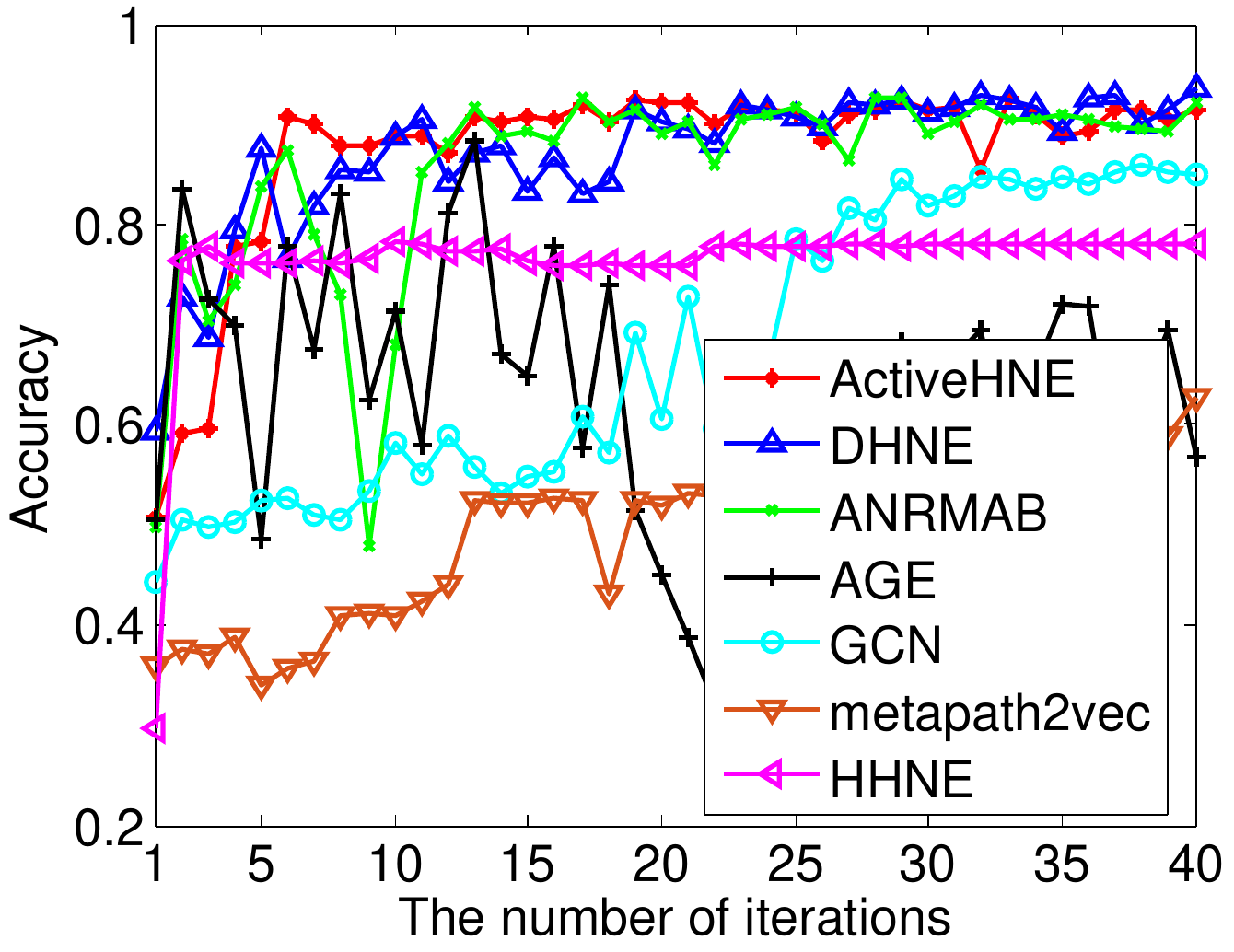}}
\vspace{-1em}
\caption{Accuracy vs. number of iterations for all methods on the three datasets. 
}
\label{fig2}
\end{figure*}

Following the experimental settings in \cite{Kipf2016Semi}, we randomly divide the labeled nodes into three parts: the training set (25\% of the labeled nodes), the validation set (25\% of the labeled nodes for hyperparameter optimization in DHNE), and the remaining as the testing set. For AL settings, the training set is used as the unlabeled pool ($\mathcal{U}$).
All the comparing methods in AL settings iteratively query the labels of the selected batch of nodes from $\mathcal{U}$, and then add these queried nodes with labels into $\mathcal{L}$ (the set of labeled training nodes).
 For a fair comparison, we use the proposed DHNE as the basic embedding and classification method for all active learning methods (AGE and ANRMAB) in the experiments.
The non-AL methods (i.e., DHNE, GCN, metapath2vec, HHNE), randomly select the nodes to label in each iteration of AL.
To evaluate the classification performance of metapath2vec and HNNE, we train a logistic regression classifier using the respective embedding of nodes.
 ActiveHNE can work with the zero-start setting (i.e., no labeled nodes, $\mathcal{L}\subseteq \varnothing$, at the beginning of active learning) using $\phi_{nc}$.
AGE and ANRMAB can operate in the same manner as ActiveHNE.
In the following, we run each method ten times   and report the average results.
\subsection{Comparison against State-of-the-art Methods}
The goal of ActiveHNE is to improve the classification performance with fewer queried nodes as much as possible.
Figure 2 shows the accuracy of all the comparing methods on the three datasets, as a function of the number of iterations. 
One iteration corresponds to $b$ nodes.
We set the batch size $b=20$ for Cora and MovieLens, and $b=5$ for DBLP, to display the difference in accuracy with respect to the number of iterations.

From Figure \ref{fig2}, we can make the following observations:\\
(i) {\bf Active vs. naive-active}.
ActiveHNE, an active method that combines DHNE and AHQN, significantly outperforms naive-active methods (DHNE, GCN, HHNE and metapath2vec), which randomly select nodes for query. It proves that AL is conducive to improve embedding for classification purpose. \\
(ii) {\bf ActiveHNE vs. other active methods}.  ActiveHNE outperforms other AL-assisting methods (ANRMAB and AGE) on MovieLens and Cora, and has comparable performance with ANRMAB on DBLP. Since these three methods use the same embedding module and only differ on the active learning strategy, the superior performance of ActiveHNE validates the effectiveness of our designed active query strategy.  Although ANRMAB and AGE are based on the same DHNE. They lose to DHNE in most cases. That is because these methods don't consider the heterogeneity and dependency of nodes in HINs. These results demonstrate the effectiveness of our proposed AQHN for DHNE.\\
(iii) {\bf DHNE vs. other network embedding methods}. DHNE significantly outperforms the three representative network embedding methods (GCN, HHNE and metapath2vec), when they are all in naive-AL setting. This observation shows the  superiority of DHNE in embedding HINs for nodes classification, and it also justifies the rationality of dividing HINs into homologous networks and bipartite networks.
The poor performance of HHNE and metapath2vec may be caused by the improper meta-path schemes and the sensibility of parameters in metapath-based random walks.

We used the Wilcoxon signed-rank test over these three datasets to evaluate the significance by dividing the Accuracy curve of each comparing method in Figure \ref{fig2} into 40 bins (each bin corresponds to a batch of queried nodes during active learning). The obtained p-values between ActiveHNE and the comparing methods are all smaller than $5 \times 10^{-4}$, except that the $p$-values for DHNE and ANRMAB on DBLP are $0.5908$ and $0.0328$, and the $p$-value for GCN on Cora is 0.0201. These results prove the statistical significance of ActiveHNE in most cases.

\subsection{Effectiveness of Individual Selection Strategy}
\label{sec:Individual}
In Section \ref{sec:Selection Strategy}, we use three node selection strategies: NC, CIE and CID. The latter two are our proposed novel strategies. 
To validate their effectiveness, 
we introduce  five variants: 
\begin{itemize}
\item {\bf ActiveHNE-nc} only uses  NC $\phi_{nc}$;
\item {\bf ActiveHNE-cie} only uses the  CIE $\phi_{cie}$ in Eq. (\ref{eq11});
\item {\bf ActiveHNE-ie} only uses  the original information entropy $\phi_{ie}(v_i)=-\sum_{c=1}^{C}\mathbf{F}_{ic}\log \mathbf{F}_{ic}$;
\item {\bf ActiveHNE-cid} only uses CID $\phi_{cid}$ in Eq. (\ref{eq13});
\item {\bf ActiveHNE-id} only uses the original information density $\phi_{id}(v_i)=\frac{1}{1+dis(\mathbf{E}_i,\varphi(v_i))}$.
\end{itemize}

\begin{figure}[ht!bp]
\centering
\hspace{-1em}
\subfigure[MovieLens]{\label{fig3a}\includegraphics[scale=0.3]{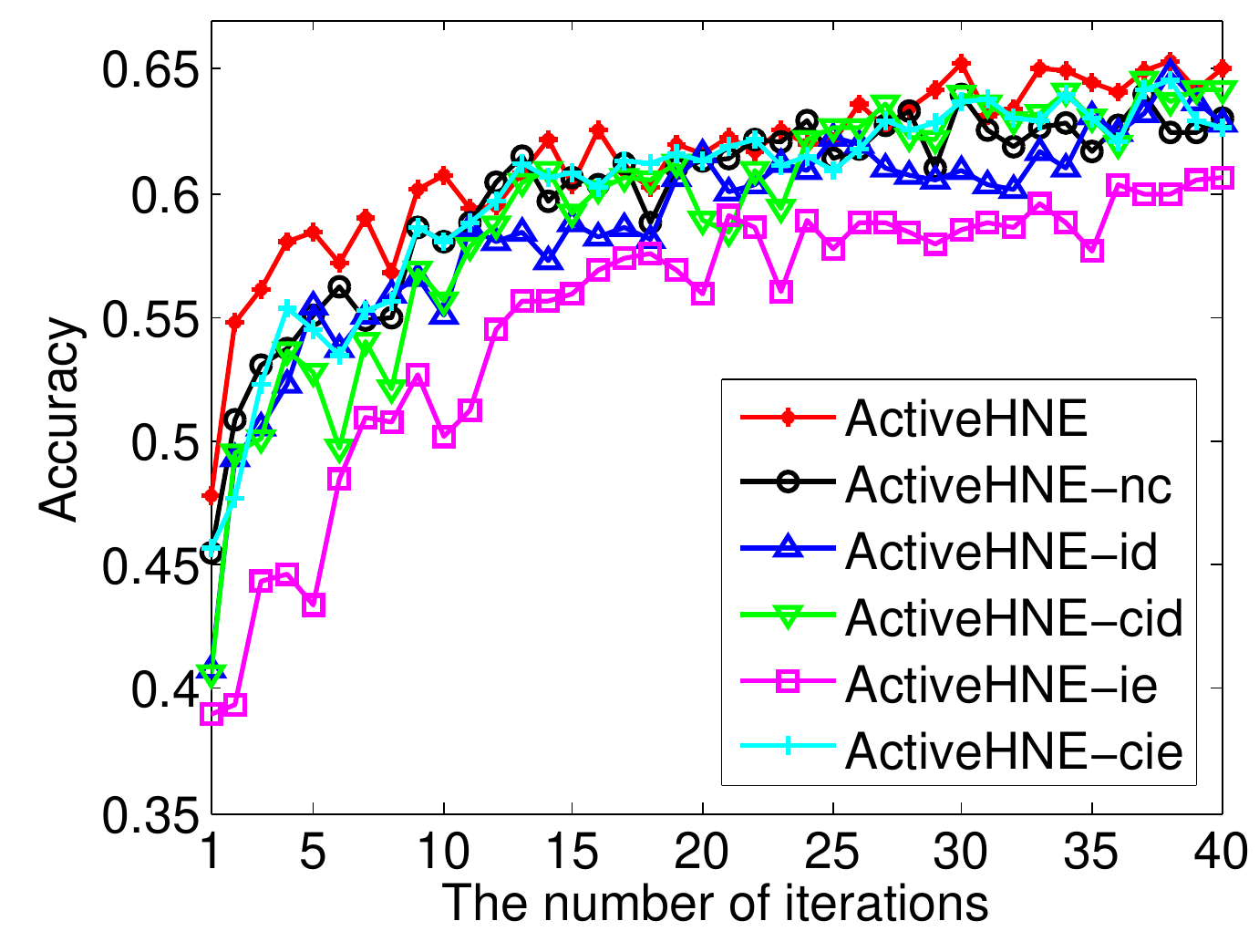}}
\subfigure[Cora]{\label{fig3b}\includegraphics[scale=0.3]{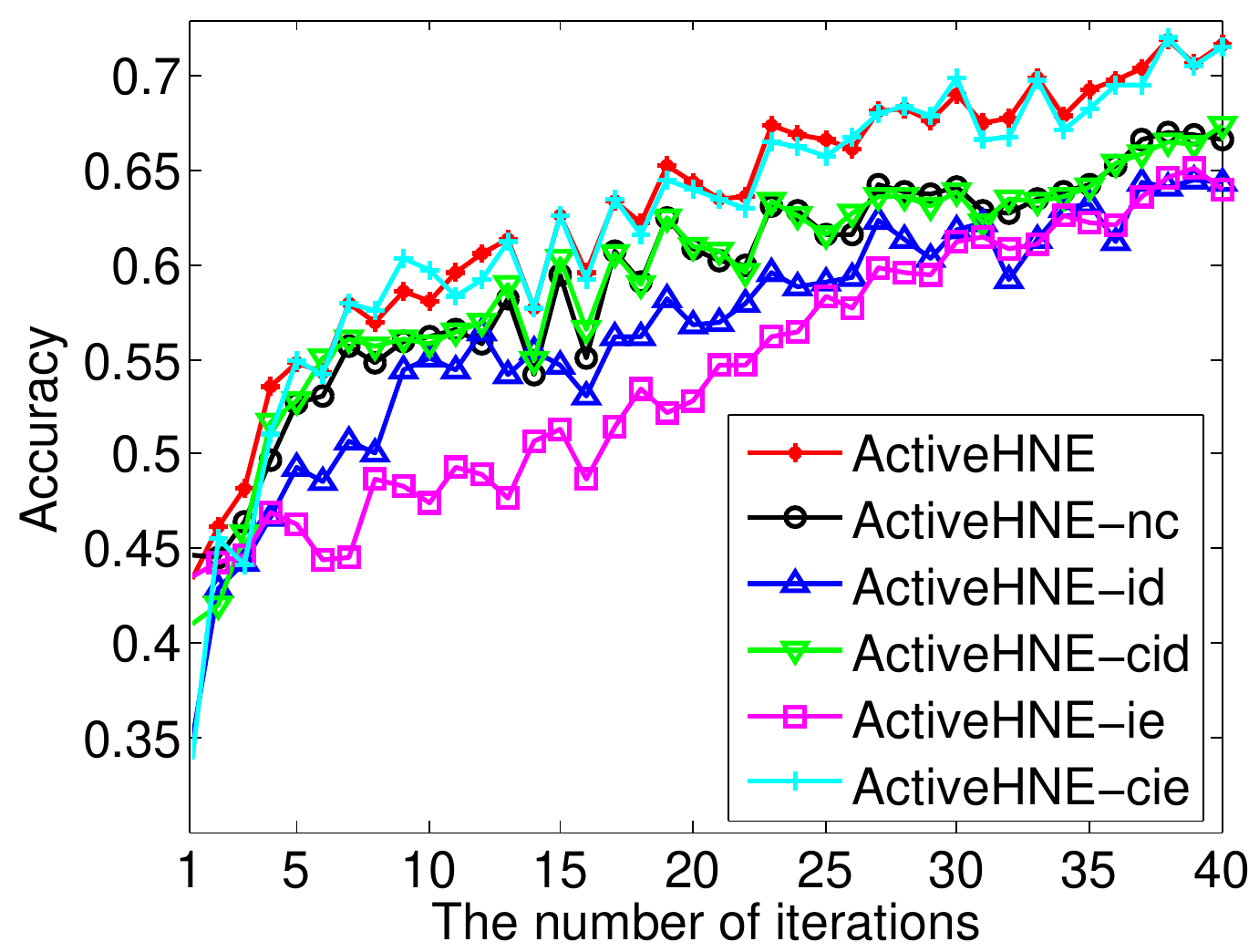}}
\vspace{-0.5em}
\caption{Accuracy vs. number of iterations: ActiveHNE against its four variants on MovieLens and Cora.}
\label{fig3}
\end{figure}

The same settings in Figure \ref{fig2} are used, and results are shown in Figure \ref{fig3}.
From Figure \ref{fig3}, we can conclude the following: \\
(i) {\bf ActiveHNE achieves the best} accuracy among its variants. Although ActiveHNE-cie also obtains the comparable accuracy to ActiveHNE on Cora, it significantly loses to ActiveHNE on MovieLens. These results support the rationality and effectiveness of ActiveHNE in combining three active selection strategies, since one particular strategy cannot fit all datasets. \\
(ii) {\bf ActiveHNE-cie and ActiveHNE-cid }
achieve a { \bf better} accuracy than {\bf ActiveHNE-ie and ActiveHNE-id}, respectively. This result corroborates the effectiveness of our proposed CIE and CID in selecting the most uncertain nodes and most representative nodes.

Likewise, we performed the Wilcoxon signed-rank test to assess the significance between ActiveHNE and the other five variants in Figure \ref{fig3}. The $p$-values for ActiveHNE with respect to all comparing methods on MovieLens and Cora are all smaller than $3\times10^{-6}$, except that the $p$-value for ActiveHNE with respect to ActiveHNE-cie on Cora is $0.0044$. All these $p$-values show the superiority of ActiveHNE.

\begin{figure}[h!t]
\centering
\hspace{-1em}
\subfigure[MovieLens]{\label{fig4a}\includegraphics[scale=0.3]{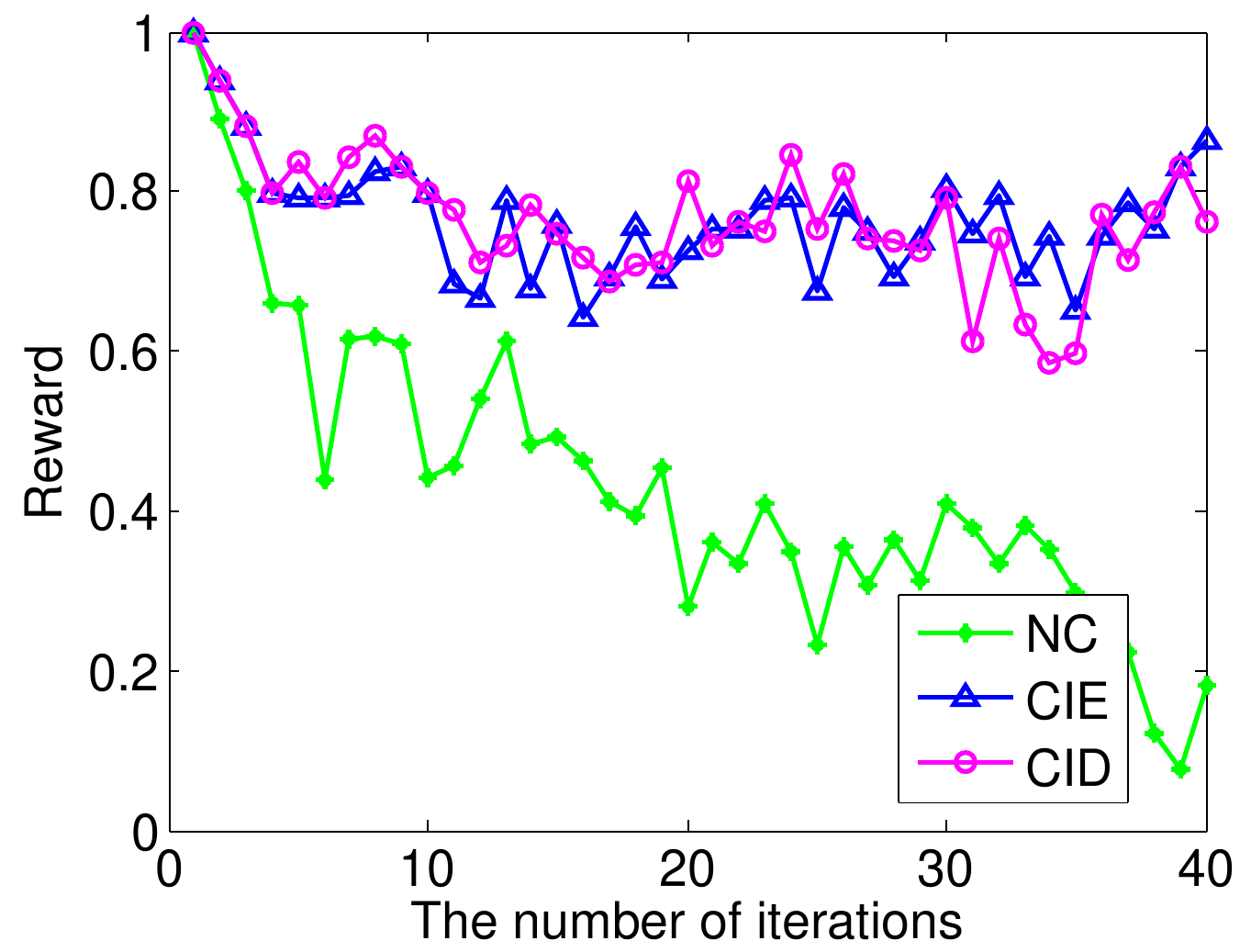}}
\subfigure[Cora]{\label{fig4b}\includegraphics[scale=0.3]{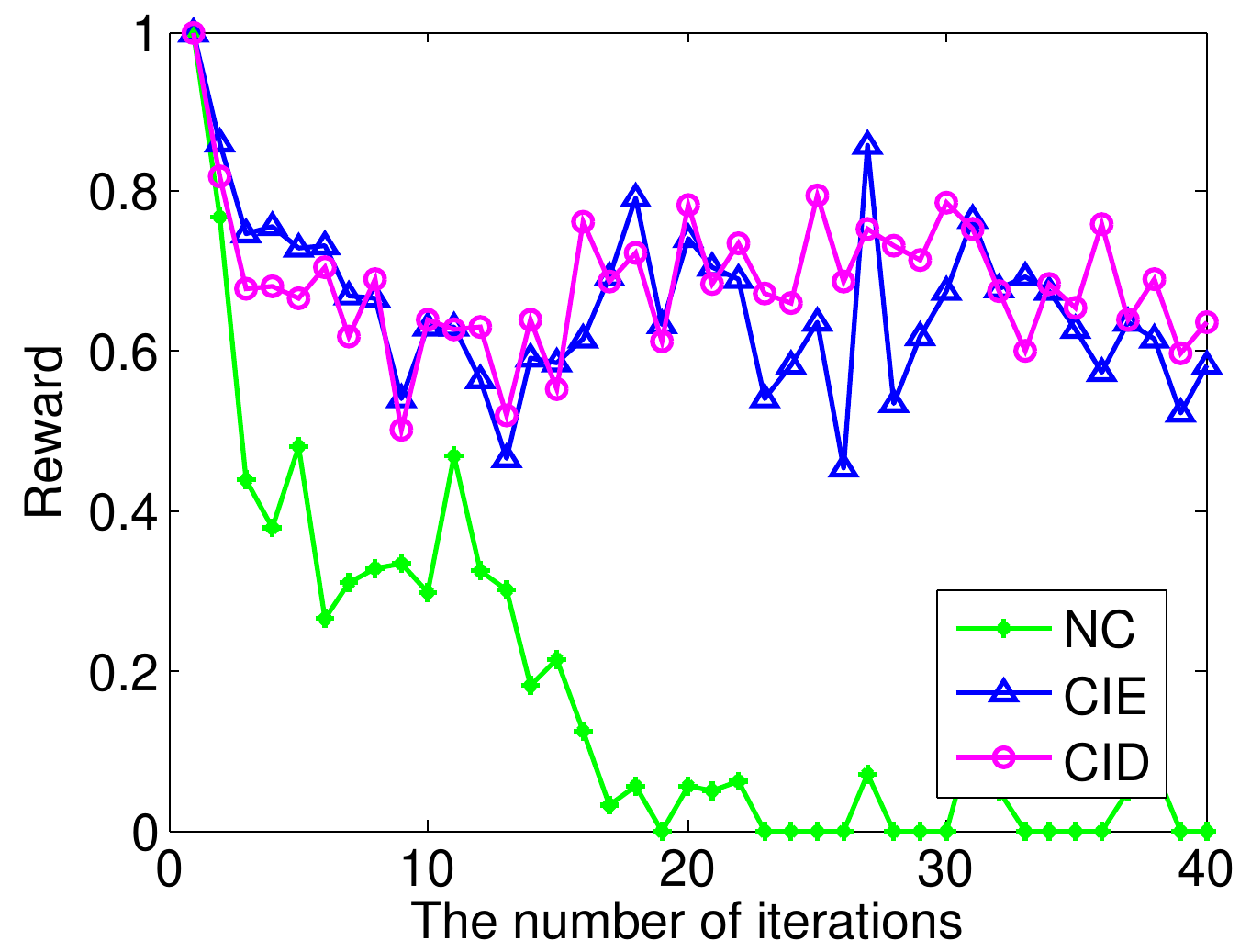}}
\vspace{-0.5em}
\caption{Reward vs. number of iterations of the three selection strategies on MovieLens and Cora. }
\label{fig4}
\end{figure}

To intuitively show the importance of each selection strategy during the AL process, we report the reward changes of different strategies in Figure \ref{fig4}, where the initial reward of each strategy is one.
From Figure \ref{fig4}, we can  see that the reward of NC gradually reduces as the number of iterations increases.
The reason is that the embedding model doesn't perform well in the initial stage of AL because of scarce labels, and CIE and CID depend on the outputs of the embedding model while NC does not. At the beginning, NC contributes to reduce the effect of the bias induced by CIE and CID. 
As the number of iterations increases, CIE and CID become more reasonable and thus the importance of NC decreases.
We observe that the sum of the rewards of NC, CIE, and CID is greater than or equal to one. This is because the nodes selected by the strategies overlap. Thus, the fact that the reward of NC goes down doesn't mean that the rewards of CIE and CID go up.
\vspace{-1cm}
\vspace{-1em}
\subsection{Parameter sensitivity analysis of $K$}
\label{parameter_k}
\begin{figure}[t]
\centering
\includegraphics[width=6cm,height=4cm]{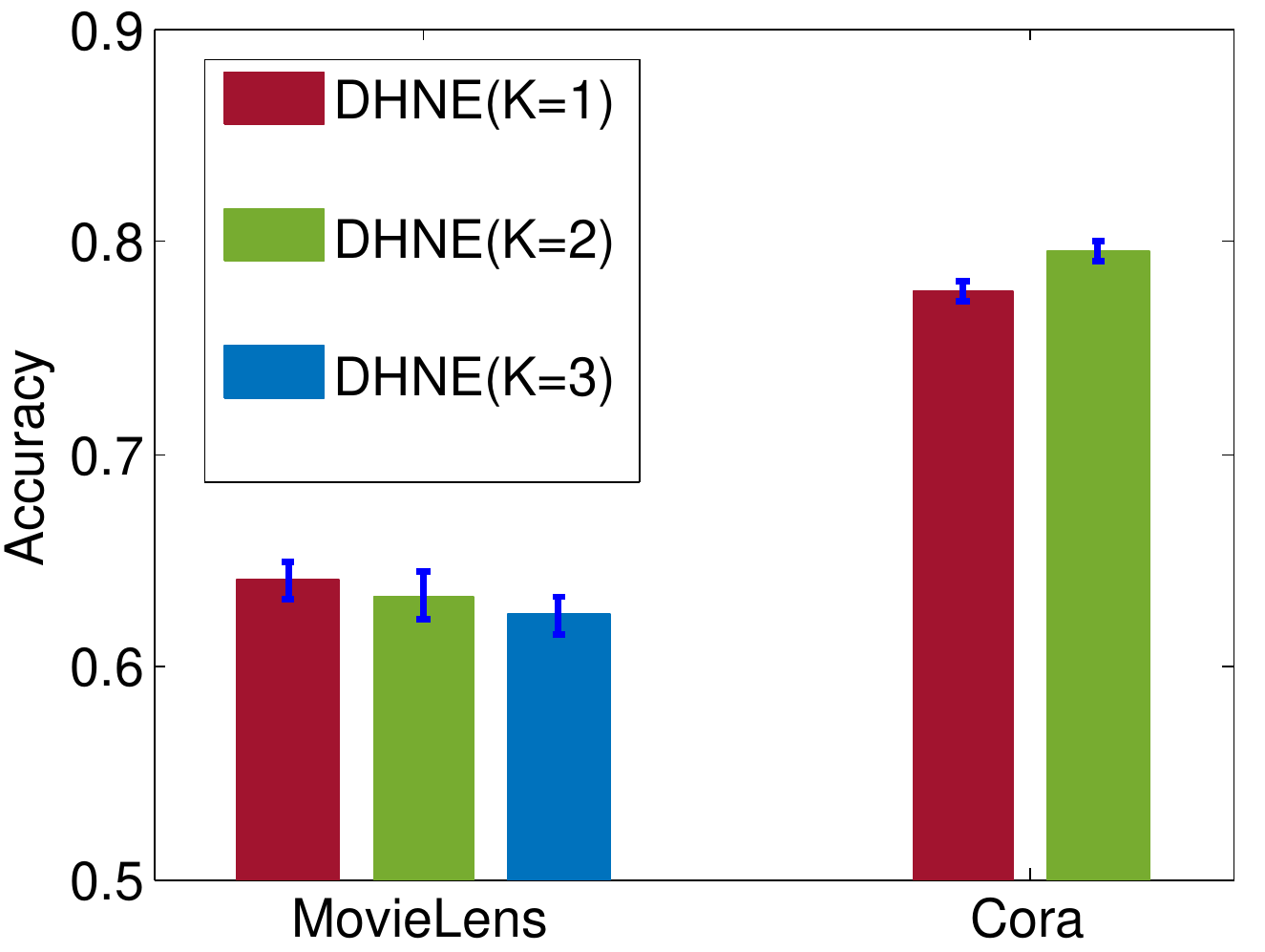}
\vspace{-0.5em}
\caption{Accuracy of ActiveHNE under different values of K on MovieLens and Cora datasets.}
\label{fig5}
\end{figure}

In the previous experimental setup, we simply set $K=1$ for DHNE and ActiveHNE. It means that DHNE and ActiveHNE only consider the one-order neighborhood structures of networks. In view of the importance of high-order neighborhood structures in real-world networks, we equip DHNE with different $K$ values in the range $\{1,2,3\}$ to investigate the importance of order of neighborhood structure, and report the average \emph{Accuracy} of 10 independent runs in Figure \ref{fig5}. 
Particularly, Figure \ref{fig5} only reports the results of DHNE on MovieLens (with $K=1,2,3$) and Cora (with $K=1,2$). The reason is that a larger $K$ gives a more dense matrix $\mathbf{P}^{K}$, which requires more space and computation, and results in running out of memory.

From Figure \ref{fig5}, we can observe the following: i) On MovieLens, DHNE achieves the best classification performance with $K=1$, and has reduced performance as $K$ increases; ii) On Cora, DHNE($K=2$) outperforms DHNE($K=1$). Thus, we can conclude that the optimal $K$ values for DHNE on different datasets are difficult to set uniformly. Still, DHNE can achieve a good classification performance by simply setting $K=1$ from Figure \ref{fig2}.

\subsection{Runtime Analysis}
ActiveHNE consists of two components, DHNE and AQHN. We separately compare the runtimes of DHNE and AQHN with those of the other methods (GNC, metapath2vec, and HHNE for network embedding; AGE and ANRMAB for active learning).
Table 1 reports the empirical runtimes of the methods on a server with  Intel Xeon E5-2678v3, 256GB RAM, and Ubuntu 16.04.4. 
 Table \ref{table1} shows: \\
 -- for the NE part, DHNE  is  slower than GCN, but significantly faster than metapath2vec and HHNE. DHNE divides the original HINs into multiple sub-networks, and trains the model parameters on each sub-network. As such, it optimizes more parameters than GCN, despite the fact that both use graph convolution-based network embedding.\\ 
-- for the AL part, we admit that AQHN is slower than AGE and ANRMAB, but actively acquired more effective nodes for achieving better embedding results, as shown in Figure \ref{fig2}.
\vspace{-0.5em}
\begin{table}[h!t]
\scriptsize										
\centering							
\caption{Runtimes  (in seconds).}
\vspace{-0.5em}							
\label{table1}							
\begin{tabular}{l l r r r r}							
\hline	
 & & MovieLens & Cora & DBLP & Total\\
\hline
\multirow{4}{*}{NE}
& DHNE	 & $92$	 & $199$	 & $147$	 & $438$\\
& GCN	 & $38$	 & $148$	 & $97$	 & $283$\\
& metapath2vec	 & $172$	 & $720$	 & $694$	 & $1586$\\
& HHNE	 & $893$	 & $13081$	 & $3180$	 & $17154$\\
\hline
\multirow{3}{*}{AL}
& AQHN	 & $8$	 & $54$	 & $13$	 & $75$\\
& AGE	 & $2$ & $5$ & $3$	 & $10$\\
& ANRMAB	 & $1$ & $6$ & $2$	 & $9$\\
\hline		
\end{tabular}							
\end{table}
\vspace{-1em}
\section{Conclusion}
In this paper, we studied how to achieve active discriminative heterogeneous network embedding by optimally acquiring and using labels of network nodes. The proposed   framework   ActiveHNE extends graph convolution networks to heterogeneous networks by dividing the given network into multiple homogeneous and bipartite sub-networks, and performing convolutions on these networks.  Three different query strategies based on convolutions are combined to query the labels of the most valuable nodes, which are fed back for the next round of discriminative network embedding.
ActiveHNE achieves a superior or comparable performance to other methods both in terms of accuracy and efficiency.
The code and the supplemental file of ActiveHNE will be made publicly available.
\appendix

\bibliographystyle{named}
\bibliography{ActiveHNE_Bib}

\end{document}